%% file: main.tex
\documentclass[letterpaper]{article} 
\usepackage{aaai2026}  
\usepackage{times}  
\usepackage{helvet}  
\usepackage{courier}  
\usepackage[hyphens]{url}  
\usepackage{graphicx} 
\usepackage{natbib}  
\usepackage{caption} 
\usepackage{algorithm}
\usepackage{algorithmic}
\input{figures/colors}
\usepackage{url}
\usepackage{xspace}
\usepackage{listings}
\usepackage{booktabs}
\usepackage{subfigure}
\usepackage{enumitem}
\usepackage{multirow}
\usepackage{multicol}
\usepackage{xcolor}
\usepackage{tcolorbox}
\usepackage{arydshln}
\usepackage{amssymb}
\usepackage{color}
\usepackage{comment}
\usepackage{pgfplots}
\pgfplotsset{compat=1.18}
\usepgfplotslibrary{external}
\usepackage{pifont}
\usepackage{amsmath}

\usepackage{newfloat}
\usepackage{listings}
\DeclareCaptionStyle{ruled}{labelfont=normalfont,labelsep=colon,strut=off} 
\floatstyle{ruled}
\newfloat{listing}{tb}{lst}{}
\floatname{listing}{Listing}
\title{Efficient and Adaptive Simultaneous Speech Translation\\with Fully Unidirectional Architecture}

\author{Biao Fu$^{1,5,}$\thanks{\,\, These authors contributed equally to this work.}\thanks{\,\, Work done during internship at Tongyi Lab.}, 
Donglei Yu$^{2,}$\footnotemark[1]\footnotemark[2], 
Minpeng Liao$^{3,}$\thanks{\,\, Corresponding author.}, 
Chengxi Li$^{3}$, Xinjie Chen$^{3,4}$\\
\textbf{Yidong Chen}$^{1,5}$,
\textbf{Kai Fan}$^{3,}$\footnotemark[3],
\textbf{Xiaodong Shi}$^{1,5,}$\footnotemark[3]
}
\affiliations{
$^{1}$ School of Informatics, Xiamen University\\
$^{2}$ University of Chinese Academy of Sciences\\
$^{3}$ Alibaba Token Hub, Alibaba Group\\
$^{4}$ Zhejiang University\\
$^{5}$ Key Laboratory of Digital Protection and Intelligent Processing of Intangible Cultural Heritage of Fujian and Taiwan (Xiamen University), Ministry of Culture and Tourism\\
biaofu@stu.xmu.edu.cn, \{yudonglei.ydl, minpeng.lmp, chenxinjie.cxj\}@alibaba-inc.com
}
\usepackage{bibentry}

\begin{document}

\maketitle

\begin{abstract}
Simultaneous speech translation (SimulST) produces translations incrementally while processing partial speech input. 
Although large language models (LLMs) have shown strong capabilities in offline translation tasks, applying them to SimulST poses notable challenges. 
Existing LLM-based SimulST approaches either incur significant computational overhead due to repeated encoding of bidirectional speech encoder, or they depend on a fixed read/write policy, limiting the efficiency and performance.
In this work, we introduce Efficient and Adaptive Simultaneous Speech Translation (EASiST) with fully unidirectional architecture, including both speech encoder and LLM. 
EASiST includes a multi-latency data curation strategy to generate semantically aligned SimulST training samples and redefines SimulST as an interleaved generation task with explicit read/write tokens. 
To facilitate adaptive inference, we incorporate a lightweight policy head that dynamically predicts read/write actions. 
Additionally, we employ a multi-stage training strategy to align speech-text modalities and optimize both translation and policy behavior. 
Experiments on both in-domain (MuST-C) and out-of-domain (Europarl-ST) En-De and En-Es datasets demonstrate that EASiST offers superior latency-quality trade-offs compared to several strong baselines.
\end{abstract}

\begin{links}
\link{Code}{https://github.com/biaofuxmu/EASiST}
\end{links}

\input{sec1_introduction}

\input{sec2_related_work}

\input{sec3_method}

\input{sec4_experiments}

\input{sec5_conclusion}

\section*{Acknowledgements}
This work is supported by the National Science and Technology Major Project (Grant No. 2022ZD0116101),
the National Natural Science Foundation of China (NSFC) under Grant No. 62206295, 
the Major Scientific Research Project of the State Language Commission in the 13th Five-Year Plan (Grant No. WT135-38), the public technology service platform project of Xiamen City (No. 3502Z20231043), and Alibaba Research Intern Program.

\bibliography{anthology,custom}

\end{document}

%% file: figures/colors.tex
\definecolor{color1}{HTML}{1f77b4}
\definecolor{color2}{HTML}{ff7f0e}
\definecolor{color3}{HTML}{2ca02c}
\definecolor{color4}{HTML}{d62728}
\definecolor{color5}{HTML}{9467bd}
\definecolor{color6}{HTML}{8c564b}
\definecolor{color7}{HTML}{e377c2}
\definecolor{color8}{HTML}{7f7f7f}
\definecolor{color9}{HTML}{bcbd22}
\definecolor{color10}{HTML}{17becf}

\definecolor{skyblue}{HTML}{4FC3F7}       
\definecolor{honeyorange}{HTML}{FFB74D}   
\definecolor{mintgreen}{HTML}{81C784}     
\definecolor{rosepink}{HTML}{F48FB1}      
\definecolor{cyanblue}{HTML}{00BCD4}      
\definecolor{applegreen}{HTML}{66BB6A}    
\definecolor{orchidpink}{HTML}{EC407A}    
\definecolor{sunfloweryellow}{HTML}{FFD54F} 
\definecolor{pastelcoral}{HTML}{FF8A65}   
\definecolor{softlime}{HTML}{C0CA33}      

\definecolor{indigoblue}{HTML}{5C6BC0}
\definecolor{goldyellow}{HTML}{FFEE58}
\definecolor{coralred}{HTML}{EF5350}
\definecolor{lavenderpurple}{HTML}{BA68C8}
\definecolor{vividmagenta}{HTML}{E91E63}
\definecolor{punchpink}{HTML}{FF6F61}

%% file: sec1_introduction.tex
\section{Introduction}
\label{sec:intro}

Simultaneous speech translation (SimulST) aims to enable seamless cross-lingual communication in streaming scenarios such as live broadcasts and international meetings~\citep{ma-etal-2020-simulmt,fu-etal-2023-adapting,chen2024divergence}.
Unlike offline speech translation (ST), which relies on complete utterances before generating translations, SimulST systems must generate output incrementally while receiving partial speech input, thus requiring models to balance translation quality against latency.

Recent advances in offline ST have demonstrated that employing large language models (LLMs) as backbone architectures leads to substantial performance improvements~\cite{huang2023speech,chen-etal-2024-llast}.
However, extending LLMs to SimulST remains non-trivial.
Early efforts typically adopt a prompt-based cascaded architecture~\cite{koshkin-etal-2024-llms,koshkin-etal-2024-transllama}, where an offline ASR model (\emph{e.g.}, Whisper~\cite{radford2023robust}) transcribes streaming speech into text, which is then incrementally fed into an LLM as prompts for translation.
This pipeline, however, introduces error propagation from ASR result and additional latency for its repeated encoding of historical inputs.
To address these limitations, recent work has explored LLM-based end-to-end SimulST frameworks. 
FASST~\citep{ouyang2024fasst} introduces an attention masking strategy during training to simulate streaming conditions, allowing the LLM to reuse its key-value (KV) cache during inference, but the under-trained masking strategy may compromise translation performance. 
Alternatively, InfiniSST~\citep{ouyang2025infinisst} reformulates SimulST as a multi-turn dialogue task, enabling incremental input and output processing while retaining efficient KV cache utilization.
However, its data construction relies on word alignment tools over offline data, which can result in domain mismatch with SimulST and alignment errors—ultimately leading to suboptimal performance.

In this work, we propose \textbf{EASiST}, a novel framework for \textbf{E}fficient and \textbf{A}daptive \textbf{Si}multaneous \textbf{S}peech \textbf{T}ranslation with LLMs and unidirectional speech encoder. 
Unlike cascaded pipelines, EASiST adopts an end-to-end architecture that incrementally encodes speech through a streaming encoder and prompts an LLM to generate translations. 
To enable EASiST to effectively perform SimulST task, we curate SimulST training data by segmenting offline ST corpora into semantically aligned chunks under multiple latency settings, and reformat them into interleaved input-output sequences with explicit read/write tokens.
Moreover, we introduce a lightweight policy head that dynamically predicts read/write actions based on the LLM's hidden representations.

To stabilize the training, we propose a multi-stage training strategy that first teaches the LLM source-target interleaved translation format via text-only MT pre-training, then aligns speech and text modalities via offline ST training, and finally jointly optimizes translation and policy through multi-task supervised fine-tuning (SFT). 
During inference, EASiST employs an adaptive read-write policy that aligns with its SFT recipe while leveraging KV cache in both the streaming encoder and LLM for efficient decoding eliminating recomputation and reducing inference latency.

Our main contributions are listed as follows:

\begin{itemize}

\item We propose \textbf{EASiST}, an end-to-end framework for adaptive and efficient SimulST with with fully unidirectional architecture, allowing fully reusable cache.

\item We design a SimulST data curation pipeline to produce chunk-level monotonic pairs tailored to SimulST needs.

\item We introduce a policy module to predict read/write actions and a multi-stage training pipline that progressively learns translation format, modality alignment, and adaptive policy.  

\item Experimental results demonstrate that EASiST outperform multiple strong baselines in balancing between translation quality and latency. 
\end{itemize}

%% file: sec2_related_work.tex
\section{Related Work}
\subsection{Traditional SimulST}
SimulST generates translation before receiving the full source utterance, and typically relies on either fixed or adaptive read/write policies.
Fixed policies are primarily based on pre-defined rules, such as emitting one target word per fixed-length speech segment~\cite{ma-etal-2020-simulmt}, or applying wait-$k$ after word boundary detection~\cite{ren-etal-2020-simulspeech,zeng-etal-2021-realtrans,dong-etal-2022-learning,fu-etal-2023-adapting,zhang-etal-2023-training,zhang-feng-2023-end}.
In contrast, adaptive policies determine actions based on context, leveraging techniques such as data-driven learning~\cite{zhang-etal-2022-learning},  information transport theory~\cite{zhang-feng-2022-information}, attention-based alignment~\cite{papi-etal-2023-attention,papi2023alignatt}, divergence-guided decisions~\cite{chen2024divergence}, and transducer-based architectures~\cite{liu-etal-2021-cross,tang-etal-2023-hybrid}.
In addition, offline-trained ST models and pretrained encoders have been adopted to enhance SimulST performance~\cite{zhang-etal-2023-training,fu-etal-2023-adapting,fu-etal-2024-wav2vec}.

\subsection{LLM-based SimulST}
Recent advances in LLMs have led to a paradigm shift in the MT area~\cite{xu2024a,huang2023speech}, 
including simultaneous MT (SimulMT)—often used as a module in cascaded SimulST systems.
Several methods incrementally update the prompts under the fixed policies (\emph{e.g.}, wait-$k$)~\cite{wang2023simultaneous,koshkin-etal-2024-llms,koshkin-etal-2024-transllama,agostinelli-etal-2024-simul} or by integrating a traditional adaptive SimulMT model as the policy module~\cite{guo2024agent}, but suffer from recomputation due to KV cache invalidation.
To enable cache reuse, SimulMask~\cite{raffel-etal-2024-simultaneous} introduces attention masking to mimics inference behavior during training,
while ~\citet{wang2024conversational} reformulates SimulMT as multi-turn dialogue.
\citet{fu-etal-2025-llms} further enables adaptive translation with interleaved generation format and learned policies.

Recently, end-to-end LLM-based SimulST systems have been explored to avoid cascading errors and high latency.
FASST~\cite{ouyang2024fasst} introduces consistency masks, akin to SimulMask, to reduce recomputation; however, it still relies on a fixed policy and cannot adapt read/write decisions based on input semantics.
InfiniSST~\cite{ouyang2025infinisst} extends dialogue-based generation to SimulST for efficient inference, but it depends on word alignment tools over offline parallel data for data construction, which may introduce domain mismatch with SimulST and suffer from alignment errors, ultimately degrading model performance. 
In this paper, we introduce an end-to-end framework with fully unidirectional architecture for efficient and adaptive SimulST that leverages interleaved SimulST data, a lightweight policy module, and multi-stage optimization.

%% file: sec3_method.tex
\section{Methods}
\label{sec:method}

\input{figures/arch}

In this section, we present \textbf{EASiST}, a novel framework for Efficient and Adaptive Simultaneous Speech Translation with fully unidirectional architecture. 
An overview of our proposed method is depicted in Figure \ref{fig:arch}.

\subsection{SimulST Data Curation}

Ideally, training SimulST models requires explicitly curated and aligned SimulST data, where speech prefixes correspond incrementally with their target translations. 
However, collecting such datasets is challenging due to the high cost of manual annotation.
To this end, we propose to transform the offline ST dataset into a SimulST dataset based on \textbf{multi-latency chunk segmentation}. 

Given offline ST data represented as triplets ${(\mathbf{s}, \mathbf{x}, \mathbf{y})}$, where $\mathbf{s}$ denotes speech sequences, $\mathbf{x}$ and $\mathbf{y}$ represent the source transcription and target translation, respectively, we first leverage powerful LLMs (\emph{e.g.}, GPT-4) to semantically segment transcripts and generate monotonic translations under multiple latency settings, following the approach proposed in \citet{fu-etal-2025-llms}.
As illustrated in the data curation process in Figure~\ref{fig:arch}, the LLM segments the source transcription $\mathbf{x}$ into semantically independent chunks while simultaneously generating the corresponding translation chunks, which are shown in different colors.

\input{tables/data_qua}

Formally, the generated aligned SimulMT data is represented as $\mathbf{c}_\text{mt} = [(\mathbf{c}_1^x, \mathbf{c}_1^y), \cdots, (\mathbf{c}_{I}^x, \mathbf{c}_{I}^y)]$, where $\mathbf{c}_i^{[\cdot]}$ denotes the $i$-th semantic chunk in the source or target language, and $I$ is the number of chunks determined by the prompted latency requirements. 
In practice, we use three different latency settings, resulting in $I_\text{low} \geq I_\text{medium} \geq I_\text{high}$, \emph{i.e.}, three possible SimulMT pairs derived from one offline pair.
Subsequently, we employ the Montreal Forced Aligner to temporally align the source text chunks $[\mathbf{c}_1^x, \cdots, \mathbf{c}_{I}^x]$ with the original speech sequence $\mathbf{s}$, thereby obtaining the corresponding speech chunks $[\mathbf{c}_1^s, \cdots, \mathbf{c}_{I}^s]$. 
As a result, we construct well-aligned SimulST data: $\mathbf{c}_\text{st} = [(\mathbf{c}_1^s, \mathbf{c}_1^y), \cdots, (\mathbf{c}_{I}^s, \mathbf{c}_{I}^y)]$, which can be directly used to effectively train SimulST models. 

In this work, we curate a large-scale SimulST dataset based on the MuST-C corpus, containing 217K instances for En$\rightarrow$De and 294K for En$\rightarrow$Es. 
Unlike conventional offline ST data, where global reordering may result in non-monotonic alignments, our method enforces locally monotonic chunk-level alignment between speech and translation, better matching the requirements of SimulST.
To evaluate the quality of our constructed data, 
we compute the monotonicity scores based on word alignment statistics~\cite{fu-etal-2023-adapting} and report CometKiwi scores \cite{rei-etal-2022-cometkiwi-simplify}.
As shown in Table~\ref{tab:data_qua}, our data achieve significantly lower monotonicity scores while maintaining comparable or better CometKiwi scores, making it more suitable for SimulST.

\subsection{Model Architecture}

To better leverage our curated chunk-level SimulST data, we propose a fully unidirectional architecture for training an LLM-based SimulST model using an interleaved format. 

\textbf{Streaming Encoder.} 
The first unidirectional module we adopt is wav2vec-S~\cite{fu-etal-2024-wav2vec}, a streaming audio encoder adapted from wav2vec 2.0~\cite{baevski2020wav2vec2} to support incremental input. 
The original wav2vec 2.0 employs a bidirectional design that relies on future context, making it unsuitable for streaming applications. 
Additionally, wav2vec 2.0 does not reuse historical KV caches, instead requiring full recomputation for each incremental input—resulting in higher latency and reduced efficiency in real-time scenarios. 
To address these limitations, wav2vec-S introduces several key modifications: it replaces group normalization with layer normalization, adopts absolute sinusoidal positional encodings in place of convolution-based relative encodings, and uses block-wise self-attention instead of bidirectional self-attention—enabling efficient processing of streaming input. 

\textbf{Adapter.}
The adapter first uses a convolutional module to reduce the length of the speech encoder output. 
Specifically, it comprises two sequential 1D convolutional layers, each with a kernel size of 5 and a stride of 2. 
This setup reduces the length of the speech features by a factor of 4. 
To maintain streaming-friendly, the convolutional module is applied independently within each speech block, preserving the model’s unidirectional property. 
Finally, a linear layer projects the compressed speech features into the representation space of the LLM.

\textbf{Large Language Model.}
In offline ST tasks, the LLM generally receives complete speech input representations and then generates corresponding translation autoregressively. 
However, in SimulST, the LLM must translate based on partial speech input. 
To simulate the SimulST process, we reorganize the aligned SimulST data by interleaving speech and their corresponding translation chunks. 
Furthermore, to guide the model learning a read-write policy, we introduce two special tokens (\texttt{<|end-of-read|>} and \texttt{<|end-of-write|>}) that serve as explicit signals for the model to transition between reading speech and writing translations. 
Formally, given an aligned chunks sequence $\mathbf{c}_{\text{st}}$, the reorganized SimulST sequence chunks are structured as:
\begin{multline}
\mathbf{\hat{c}}_{\text{st}}=[\mathbf{c}_1^s, \texttt{<|eor|>}, \mathbf{c}_1^y, \texttt{<|eow|>}, \cdots, \\
\mathbf{c}_I^s, \texttt{<|eor|>}, \mathbf{c}_I^y,\texttt{<|eow|>} ],
\end{multline}
where the \texttt{<|eor|>} token signals the transition to writing mode for generating the translation and the \texttt{<|eow|>} token signals the model to stop translating and start reading the next speech chunk.
This formulation ensures that the LLM learns an explicit read-write policy aligned with streaming translation from our SimulST data.

\textbf{Policy Module.}
To enable an adaptive read/write policy, in parallel to the token prediction layer of LLM, we introduce an additional linear layer, which serves as a binary classifier dynamically determining whether the model should continue reading speech input or start generating translation output at each step. 
Formally, given the hidden state $h_t$ of the last layer of the LLM at timestep $t$, this module computes a read/write probability:
\begin{equation} 
p_t = \text{softmax}(\mathbf{W}_p h_t), 
\end{equation}
where $\mathbf{W}_p \in \mathcal{R}^{2 \times d}$ are learnable parameters.

\subsection{Multi-stage Training}
To stabilize the training, we propose a multi-stage progressive training framework consisting of three stages: SimulMT Pre-training, Speech-Text Modality Alignment, and Multi-task Supervised Fine-tuning (SFT).

\textbf{Stage I: SimulMT Pre-training.}
The goal of the first stage is to guide the LLM to learn the format of interleaved source and target sequences for streaming translation.
Specifically, given an interleaved SimulMT sequence $\mathbf{\hat{c}}_{\text{mt}} = [\mathbf{c}_1^x, \texttt{<|eor|>}, \mathbf{c}_1^y, \texttt{<|eow|>}, \cdots, \mathbf{c}_I^x, \texttt{<|eor|>}, \mathbf{c}_I^y, \\ \texttt{<|eow|>}]$, the training objective for the SimulMT task is formulated as the autoregressive prediction:
\begin{equation} 
\mathcal{L}_{\text{SimulMT}} =- \sum_{t=1}^{|\mathbf{\hat{c}}_{\text{mt}}|} \log p_{\theta}(\hat{y}_t | \mathbf{o}, \hat{y}_{\leq t-1}), 
\end{equation}
where $\hat{y}_t$ represents the $t$-th token in the sequence $\mathbf{\hat{c}}_{\text{mt}}$, $\mathbf{o}$ is the prompt, and $\theta$ denotes the LLM’s parameters. In this task, we compute the cross-entropy loss for all tokens, including source text, target text, and special tokens.

Additionally, to maintain the LLM’s capability for full-sentence translation, we include an offline MT objective at this stage. 
Given a sentence pair $(\mathbf{x}, \mathbf{y})$, its loss is defined as:
\begin{equation} 
\mathcal{L}_{\text{MT}}=- \sum_{t=1}^{|\mathbf{y}|} \log p_{\theta}(y_t | \mathbf{x}, y_{\leq t-1}),
\end{equation}
where $y_t$ represents the $t$-th token in the target translation $\mathbf{y}$. 
The overall loss for this stage is then computed as:
\begin{equation} 
\mathcal{L}_{\text{Stage-I}} = \mathcal{L}_{\text{SimulMT}} + \mathcal{L}_{\text{MT}}. 
\end{equation}
At this stage, we employ full-parameter fine-tuning to train the LLM for one epoch to effectively learn the SimulMT in the novel autoregressive and interleaved format.

\textbf{Stage II: Speech-Text Modality Alignment.}
The second stage aims to align the speech and text modalities at the semantic level by training the model on an offline ST task. 
Given a parallel speech-text pair $(\mathbf{s}, \mathbf{y})$, the training loss is formulated as:
\begin{equation} 
\mathcal{L}_{\text{Stage-II}}=\mathcal{L}_{\text{ST}}=- \sum_{t=1}^{|\mathbf{y}|} \log p_{\phi}(y_t | \mathbf{s}, y_{\leq t-1}).
\end{equation}
At this stage, we freeze the LLM and train the streaming encoder and adapter. 
This ensures that the speech encoder aligns well with the LLM’s text representation space while maintaining the LLM’s translation capabilities.

\textbf{Stage III: Multi-task SFT.}
In the final stage, we enhance the model’s streaming translation capability by jointly optimizing multiple tasks, including:

(1) \textit{SimulST task}: The primary objective is to improve streaming translation performance by fine-tuning the model on the structured SimulST data. 
Given a SimulST data $\mathbf{\hat{c}}_{\text{st}}$, we define its loss as:
\begin{equation} 
\mathcal{L}_{\text{SimulST}} = -\sum_{i=1}^{I}\sum_{t=1}^{|\mathbf{c}_i^y|} \log p_{\theta}(c_{i,t}^y | \mathbf{c}_{1:i}^s, \mathbf{c}_{1:i-1}^y, c_{i, <t}^y),
\label{eq:simulst}
\end{equation}
where $\mathbf{c}_{1:i}^{[\cdot]}$ denotes the first $i$ speech or translation chunks and $c_{i,t}^y$ is the $t$-th token in the $i$-th translation chunk.
We omit the loss computation for special tokens in Eq. (\ref{eq:simulst}) without hurting readability. 
In practice, the loss is calculated on both the target text and special tokens.

(2) \textit{Policy decision task}: To train the policy module, we use a binary classification objective that guides the model in making read/write decisions, where the loss is simply a binary cross-entropy loss.
\begin{equation} 
\mathcal{L}_{\text{policy}} = -\sum_{t=1}^{T} \left[ y_t \log p_t + (1 - y_t) \log (1 - p_t) \right], 
\end{equation} 
where $y_t \in \{0,1\}$ is the ground-truth read ($0$) or write ($1$) label derived from the aligned SimulST dataset. 
Concretely, given a speech chunk $\mathbf{c}_i^s$ that is segmented into $n$ streaming \textit{blocks}, we assign decision labels based on block boundaries.
We assign read labels ($y_t=0$) to the last position of the first $n-1$ blocks, indicating that the model should continue reading. 
For the last (i.e., $n$-th) \textit{block}, we assign $y_t = 1$ (write decision) at its final position, signaling that the model should begin generating the corresponding translation.
To address the potential label imbalance between read and write decisions, we additionally assign the label $y_t = 1$ to all tokens within the corresponding translation chunk $\mathbf{c}_i^y$. 
Finally, we compute the policy loss only at the labeled positions: the final positions of speech blocks and the translation tokens. 
All other positions within the speech chunk are excluded from the loss computation.

(3) \textit{Offline ST task}: To preserve the model’s translation accuracy, we keep the offline ST loss $\mathcal{L}_{\text{ST}}$ as a regularization. 
Thus, the final objective function for this stage is:
\begin{equation} 
\mathcal{L}_{\text{Stage-III}} = \mathcal{L}_{\text{SimulST}} + \mathcal{L}_{\text{ST}} + \lambda \mathcal{L}_{\text{policy}},
\end{equation}
where $\lambda$ is a hyper-parameter controlling the weight of the policy loss. 
We set $\lambda=1$ in experiments.

In this stage, we keep the LLM frozen and fine-tune the rest of the model components, including the streaming encoder, adapter, and policy module. 
Through this multi-stage training pipeline, our model effectively learns to generate translations based on partial speech input and adaptively make read/write decisions, ultimately achieving efficient and adaptive SimulST.

\input{figures/results_simulst}

\subsection{SimulST Inference}

\textbf{Gap-Free between Train and Infer} Our model performs autoregressive inference in a manner consistent with its training process. 
Specifically, when a speech block is received, the hidden state at its final position is passed to the policy module to compute a read/write probability $p_t$. 
If the predicted probability exceeds a pre-defined threshold $\tau$, the model stops reading further speech input, appends the \texttt{<|end-of-read|>} token, and then begins generating translation tokens autoregressively until the \texttt{<|end-of-write|>} token is emitted. 
Otherwise, it continues to read the next speech block before making another decision. 
The threshold $\tau$ also serves as a tunable hyperparameter to control latency: a lower value  prompts earlier output translation, while a higher value  encourages the model to wait for more speech input.

\textbf{Reusable Cache} Our model achieves efficient streaming translation by leveraging the KV cache mechanisms of both the streaming encoder and the LLM.
Since past context is cached, the model avoids redundant computation for previously seen tokens, significantly improving computational efficiency and reducing inference latency.

%% file: figures/arch.tex
\begin{figure*}[t]
\centering
\includegraphics[width=1.0\textwidth]{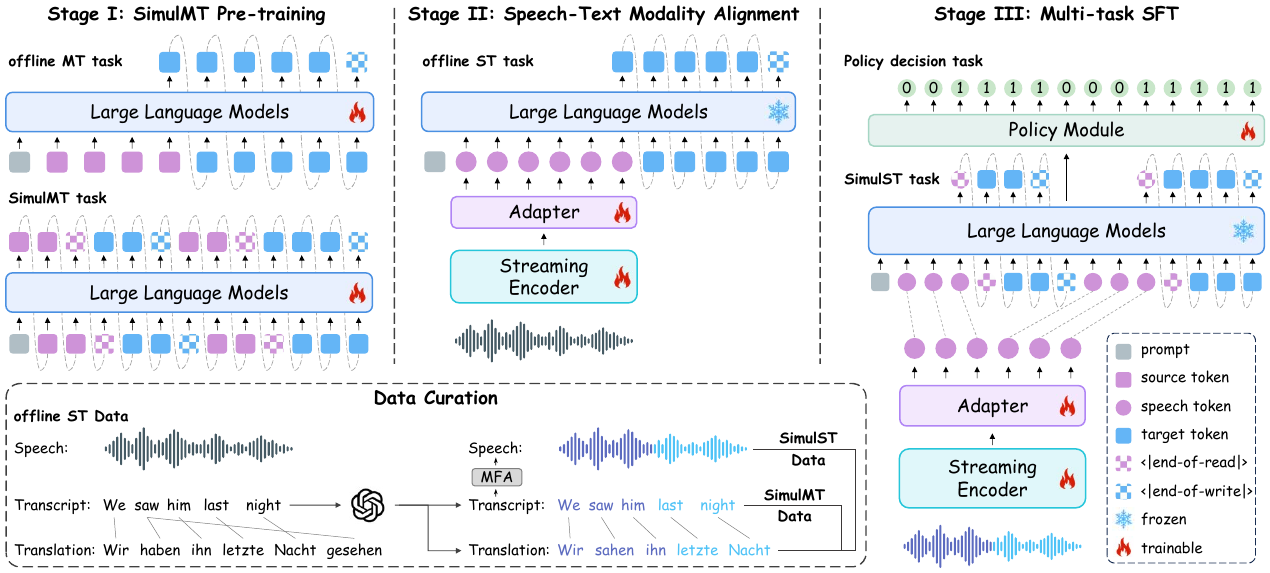}
\caption{Overview of the proposed EASiST framework. Bottom: SimulST data curation pipeline that generates monotonic interleaved SimulST data from offline ST corpora. Top: A three-stage training strategy—(I) MT pre-training on SimulMT and offline MT data, (II) speech-text modality alignment via offline ST task, and (III) multi-task SFT for optimizing SimulST and adaptive read/write policy.}

\label{fig:arch}
\end{figure*}

%% file: tables/data_qua.tex
\begin{table}[t]
\centering
\small
\begin{tabular}{lcccc}
\toprule
\multirow{2}{*}{Data} & \multicolumn{2}{c}{En$\rightarrow$De} & \multicolumn{2}{c}{En$\rightarrow$Es} \\ \cmidrule(lr){2-3} \cmidrule(lr){4-5}
 & Mono. ($\downarrow$) & Kiwi ($\uparrow$) & Mono. ($\downarrow$) & Kiwi ($\uparrow$) \\
 \midrule
Ours & 1.01 & 85.04 & 1.06 & 86.21 \\
Offline & 1.54 & 84.16 & 1.51 & 85.07 \\
\bottomrule
\end{tabular}%
\caption{Monotonicity (Mono.) and CometKiwi (Kiwi) scores of our SimulST data vs. offline ST data.}
\label{tab:data_qua}
\end{table}

%% file: figures/results_simulst.tex
\pgfplotscreateplotcyclelist{easistColorCycle}{
  {orchidpink, solid,          mark=*,        mark size=1.4pt},
  {honeyorange, solid,        mark=triangle*, mark size=1.8pt},
  {applegreen, solid, mark=square*,  mark size=1.2pt},
  {cyanblue, solid,       mark=diamond*, mark size=1.8pt},
  {indigoblue, solid, mark=pentagon*, mark size=1.8pt},
}

\pgfplotsset{
  easistAxis/.style={
    width=0.295\textwidth,
    height=0.3\textwidth,
    font=\footnotesize,
    label style={font=\footnotesize},
    tick label style={font=\footnotesize},
    legend style={
      font=\scriptsize,
      at={(0.99,0.01)},
      anchor=south east,
      legend columns=1
    },
    legend cell align={left},
    cycle list name=easistColorCycle,
    every axis plot/.append style={line width=1.0pt},
    xmajorgrids=true,
    ymajorgrids=true,
    grid style=dashed,
    enlargelimits=0.04,
    every axis y label/.append style={at={(-0.12,0.5)}},
    every axis/.append style={line width=1.0pt},
  }
}

\begin{figure*}[t]
\centering

\subfigure[{MuST-C En$\rightarrow$De}]{
\begin{tikzpicture}
\begin{axis}[
  easistAxis,
  xlabel=LAAL (s),
  ylabel=BLEU,
  ymin=18, ymax=30,
  xmin=1,
  ytick={18,22,26,30},
]
\addplot coordinates {(1.26,23.09)(1.64,25.5)(2.31,27.58)(3.23,28.59)(4.28,29.36)(5.01,29.77)}; 
\addplot coordinates {(1.91,17.64)(2.69,23.87)(3.39,26.03)(3.94,27.06)(4.37,27.78)(4.82,27.75)(5.1,28.51)}; 
\addplot coordinates {(1.26,21.86)(1.68,26.27)(2.26,27.62)(3.28,28.02)(4.15,28.15)(4.94,27.57)}; 
\addplot coordinates {(1.25,21.29)(1.68,24.37)(2.55,27.3)(3.27,28.73)(4.25,29.35)(5.04,29.24)}; 
\addplot coordinates {(1.07,20.94)(1.53,24.56)(2.02,25.56)(2.49,26.23)(2.96,26.19)}; 
\legend{EASiST, wait-$k$, AlignAtt, EDAtt, InfiniSST}
\end{axis}
\end{tikzpicture}
}
\hspace{-3mm}
\subfigure[{MuST-C En$\rightarrow$Es}]{
\begin{tikzpicture}
\begin{axis}[
  easistAxis,
  xlabel=LAAL (s),
  xmin=1,
  ytick={20,24,28,32},
]
\addplot coordinates {(1.2,26.61)(1.48,28.89)(1.9,30.32)(2.57,30.98)(3.57,31.86)(4.69,32.9)}; 
\addplot coordinates {(2.02,18.17)(2.79,26.81)(3.5,29.47)(4.08,30.45)(4.53,31.32)(5.03,31.93)(5.34,32.39)}; 
\addplot coordinates {(1.24,24.67)(1.64,25.81)(1.94,26.97)(2.67,30.65)(3.87,31.89)(4.8,32.95)}; 
\addplot coordinates {(1.21,25.62)(1.4,27.5)(2.0,29.4)(2.48,31.64)(3.27,33.17)(4.47,33.93)}; 
\addplot coordinates {(0.98,25.67)(1.45,29.63)(1.94,31.39)(2.43,32.23)(2.94,32.45)}; 
\legend{EASiST, wait-$k$, AlignAtt, EDAtt, InfiniSST}
\end{axis}
\end{tikzpicture}
}
\hspace{-3mm}
\subfigure[{Europarl-ST En$\rightarrow$De}]{
\begin{tikzpicture}
\begin{axis}[
  easistAxis,
  xlabel=LAAL (s),
  ymax=26,
  ytick={14,18,22,26},
]
\addplot coordinates {(1.46,18.87)(2.11,22.12)(3.38,24.01)(5.12,25.17)(6.78,25.5)(7.73,25.45)}; 
\addplot coordinates {(2.34,13.74)(3.19,22.01)(4.08,24.61)(4.92,24.61)(5.66,25.0)(6.57,25.13)(7.23,25.26)}; 
\addplot coordinates {(1.64,13.94)(2.06,20.08)(3.28,23.9)(4.95,24.93)(6.37,24.97)(7.91,25.09)}; 
\addplot coordinates {(1.79,16.73)(2.54,20.09)(3.28,22.48)(5.08,24.48)(6.79,25.01)(7.97,25.21)}; 
\addplot coordinates {(1.28,16.05)(1.7,19.35)(2.18,21.06)(2.65,22.03)(3.15,22.54)}; 
\legend{EASiST, wait-$k$, AlignAtt, EDAtt, InfiniSST}
\end{axis}
\end{tikzpicture}
}
\hspace{-3mm}
\subfigure[{Europarl-ST En$\rightarrow$Es}]{
\begin{tikzpicture}
\begin{axis}[
  easistAxis,
  xlabel=LAAL (s),
  ytick={20,24,28,32},
]
\addplot coordinates {(1.48,27.4)(1.91,30.03)(2.7,31.06)(3.99,32.34)(5.65,33.0)(7.01,33.1)}; 
\addplot coordinates {(2.61,16.38)(3.4,31.06)(4.27,32.77)(5.07,32.83)(5.79,33.02)(6.64,33.08)(7.28,33.0)}; 
\addplot coordinates {(1.68,19.23)(1.96,20.71)(2.83,24.53)(4.24,32.57)(5.58,32.91)(6.9,33.02)}; 
\addplot coordinates {(1.6,21.98)(2.08,26.49)(2.84,29.68)(4.21,32.19)(5.82,32.84)(6.84,32.93)}; 
\addplot coordinates {(1.24,22.14)(1.68,26.46)(2.13,28.48)(2.61,29.37)(3.11,29.64)}; 
\legend{EASiST, wait-$k$, AlignAtt, EDAtt, InfiniSST}
\end{axis}
\end{tikzpicture}
}

\caption{The translation quality (BLEU) against the latency metric (LAAL) on both in-domain (MuST-C En$\rightarrow$De/Es) and out-of-domain (Europarl-ST En$\rightarrow$De/Es) test sets.}
\label{fig:results_simulst}
\end{figure*}

%% file: sec4_experiments.tex
\section{Experiments}
\label{sec:experiments}

\subsection{Experimental Settings}

\textbf{Datasets.}
For the offline MT and ST tasks, we use the training sets from the MuST-C v1 \citep{di-gangi-etal-2019-must} English$\rightarrow$German (En$\rightarrow$De) and English$\rightarrow$Spanish (En$\rightarrow$Es) datasets, which contain speech, transcription, and translation triplets.
For the SimulMT and SimulST tasks, we train our models on the SimulST dataset constructed by our multi-latency segmentation method.
We evaluate our method on the \texttt{tst-COMMON} set of MuST-C and additionally assess generalization to out-of-domain settings using the Europarl-ST~\cite{iranzo2020europarlst} En$\rightarrow$De and En$\rightarrow$Es test sets.

\textbf{Implementation Details.} 
EASiST uses the finetuned wav2vec-S-Large \cite{fu-etal-2024-wav2vec} as the streaming encoder and the Llama-3-8B-Instruct  \cite{dubey2024llama} as the backbone LLM. 
The adapter consists of two 1D convolutional layers followed by a linear projection layer.  
Across all stages, we use a batch size of 128, a cosine learning-rate scheduler, and the AdamW optimizer.
Stage I trains the LLM for 1 epoch with micro-batch size 16 using learning rate 1e-5, and warmup ratio 0.1, updating 8.03B parameters. Stage II trains the encoder and adapter for 6 epochs with micro-batch size 4 using learning rate 2e-4, and warmup ratio 0.03, updating 323M parameters. Stage III uses the same setup as Stage II except training for 1 epoch with learning rate 2e-5, updating 323M parameters.
All experiments are run a single time on 8 NVIDIA A100 80G GPUs.

\textbf{Evaluation.} 
For translation quality, we report case-sensitive detokenized BLEU using SacreBLEU.  
For latency, we adopt LAAL \citep{papi-etal-2022-generation} and computational-aware LAAL (LAAL-CA). 
We evaluate EASiST under varying probability thresholds $\tau \in \{0.1, 0.2, \ldots, 0.6\}$ to control the latency-quality trade-off.
All evaluations are conducted using greedy decoding.

\textbf{Baseline Models.} 
We first implement an \textbf{offline} ST system following \citet{zhang2023tuning}, which comprises a wav2vec 2.0 encoder, a convolutional adapter, and an LLM.
The model is trained in two stages: fine-tuning the LLM on offline MT data and aligning speech and text modalities on offline ST data.
The training setup is the same as the first two phases of EASiST. 
We compare EASiST with several strong end-to-end SimulST approaches, including \textbf{wait-$k$}~\cite{ma-etal-2019-stacl}, \textbf{EDAtt} \citep{papi-etal-2023-attention}, \textbf{AlignAtt} \citep{papi2023alignatt} and InfiniSST \cite{ouyang2025infinisst}.
Except for InfiniSST, we implement all baselines on top of the offline ST model; for InfiniSST, we evaluate using its official released weights.

\input{figures/results_simulst_ca}

\subsection{Main Results}
\label{sec:main_results}
\textbf{Latency-Quality Trade-off.}
Figure~\ref{fig:results_simulst} presents BLEU-LAAL trade-offs on both in-domain (MuST-C) and out-of-domain (Europarl-ST) test sets.
On the MuST-C En$\rightarrow$De and En$\rightarrow$Es directions, EASiST consistently matches or outperforms all baselines, particularly in low-latency regions (LAAL $\approx$ 1s), where it achieves over +1 BLEU improvement.
More notably, EASiST demonstrates strong generalization to the out-of-domain Europarl-ST dataset. It surpasses all baselines across all latency levels, with over 2 BLEU gains on En$\rightarrow$De and more than 3 BLEU improvements on En$\rightarrow$Es at medium-to-low latency (LAAL $\approx$ 2s). These results highlight the effectiveness and cross-domain robustness of our approach.

\textbf{Computational-Aware Evaluation.}
We further evaluate the computational-aware latency (LAAL-CA) of different systems under a realistic hardware setting. All experiments are conducted on the same machine using a single NVIDIA A100 GPU to ensure fair comparison. 
As shown in Figure~\ref{fig:results_simulst_ca}, EASiST's advantages become even more pronounced under the more realistic LAAL-CA metric.
This is attributed to its fully unidirectional architecture, which allows for effective cache reuse in both the streaming encoder and the LLM, significantly reducing inference overhead. 
In contrast, the baselines lack this capability and suffer from repeated recomputation at each decoding step, leading to inflated latency in practice.
Consequently, the BLEU improvements of EASiST over baselines grow wider under LAAL-CA, achieving more substantial gains across all latency levels and test sets.
To further quantify inference efficiency, we report the average computation time per generated token across all latency levels in Table~\ref{tab:inference_speed}. 
EASiST achieves significantly faster inference speed (28.95 ms/token) compared to all SimulST baselines, while maintaining a decoding speed comparable to the offline system (23.25 ms/token).

\input{tables/speed}

\subsection{Ablation Study}
\label{sec:ablation}

To investigate the contribution of each component in our training pipeline, we conduct ablation studies on the En$\rightarrow$De \texttt{tst-COMMON} set. 
The results are shown in Figure~\ref{fig:ablation}.

\textbf{Effect of the training objectives in Stage III.}
In the \emph{w/o Stage III $\mathcal{L}_{\text{policy}}$} variant, we remove the policy loss and the corresponding policy module.
During inference, we instead apply a fixed policy where the model reads a fixed number of speech blocks ($k$) and then generates translations until an \texttt{<|eow|>} token is emitted.
We observe that this variant achieves similar translation quality to our model in low-latency regions.
However, as latency increases, its performance degrades significantly.
We attribute this to a mismatch between training and inference: 
during training, speech chunks are semantically aligned and vary in length, while inference uses fixed-length blocks.
This mismatch becomes more severe as latency increases (\emph{i.e.}, larger speech chunks), leading to degraded performance in higher latency regions.
These results show the superiority and robustness of our adaptive policy.
In the \emph{w/o Stage III $\mathcal{L}_{\text{ST}}$} variant, we remove the offline ST loss from Stage III. This change reduces translation quality at medium and high latencies by 1--2 BLEU, suggesting that incorporating offline ST objectives during SimulST training helps maintain translation quality.

\textbf{Influence of the multi-stage training framework.}
In the \emph{w/o Stage II} variant, we skip Stage II, where  EASiST freezes the LLM and performs offline ST training for 6 epochs.
For fairness, this variant is also trained for 6 epochs during Stage III.
However, skipping Stage II underperforms in all latency regions.
In the \emph{w/o Stage I} variant, we remove Stage I, which fine-tunes the LLM on SimulMT data for one epoch to learn the interleaved translation format.
To achieve the same objective, this variant fine-tunes all model parameters for 1 epoch during Stage III.
Results show that removing Stage I leads to degraded performance across all latency regions.
In the \emph{w/o Stage I and Stage II} variant, we train SimulST directly from scratch by fine-tuning the full model's parameters for 6 epochs.
This leads to a significant decrease in model performance.
These results suggest that it is challenging for the model to simultaneously optimize multiple objectives—interleaved generation, modality alignment, and read/write policy—in a single training stage, highlighting the effectiveness of our progressive multi-stage training.

In addition to improving translation performance, our multi-stage training strategy is also more efficient in the training cost.
As shown in Figure~\ref{tab:ablation_time}, EASiST achieves over 75\% and 25\% savings in total training time compared to one-stage and two-stage training variants, respectively. %
Notably, fine-tuning the LLM during SimulMT pretraining is significantly more efficient than doing so in the SimulST stage, as text sequences are significantly shorter than speech.

\input{figures/ablation}

\input{figures/training_time}

\subsection{Effect of Fine-Tuning Different Modules}
We further investigate the effect of fine-tuning different model components during Stage III, while keeping the policy module trainable in all settings.
As shown in Figure~\ref{fig:trainable_module}, fine-tuning both the encoder and adapter (our default setting) achieves the best performance across all latency levels, offering an better balance between performance and training efficiency.
Interestingly, fine-tuning only the adapter also achieves  comparable performance.
This makes adapter-only tuning a viable alternative in resource-constrained settings.
In contrast, fine-tuning only the encoder results in clearly inferior performance.
We attribute this to a representation mismatch between the encoder and the frozen LLM: without updating the adapter, the encoder's output distribution may not align well with the LLM's representation space.
On the other hand, updating the LLM (\texttt{ALL}, \texttt{Encoder+LLM}, \texttt{Adapter+LLM}, \texttt{LLM}) introduces higher computational cost and leads to performance degradation potentially due to overfitting.
Overall, these results show that lightweight tuning strategies—particularly encoder+adapter or adapter-only fine-tuning—can achieve optimal SimulST performance while avoiding the overhead of LLM updates.

\input{figures/results_ft_module}

\subsection{Fluency Evaluation}
We report the fluency scores of translations generated by EASiST under different decision thresholds $\tau$ in Table~\ref{tab:fluency}.
As $\tau$ increases, fluency scores steadily improve for both En$\rightarrow$De and En$\rightarrow$Es directions, indicating that allowing more input speech before generating output leads to more fluent translations.
Importantly, EASiST achieves high fluency across all latency settings, with near-offline quality.

\input{tables/fluency}

%% file: figures/results_simulst_ca.tex
\begin{figure*}[t]
\centering

\subfigure[{MuST-C En$\rightarrow$De}]{
\begin{tikzpicture}
\begin{axis}[
  easistAxis,
  xlabel=LAAL-CA (s),
  ylabel=BLEU,
  ymin=18, ymax=30,
  ytick={18,22,26,30},
]
\addplot coordinates {(1.5,23.09)(1.87,25.5)(2.57,27.58)(3.55,28.59)(4.66,29.36)(5.44,29.77)}; 
\addplot coordinates {(2.19,17.64)(2.94,23.87)(3.6,26.03)(4.11,27.06)(4.52,27.78)(4.94,27.75)(5.2,28.51)}; 
\addplot coordinates {(1.71,21.86)(2.39,26.27)(3.3,27.62)(5.11,28.02)(6.89,28.15)(8.92,27.57)}; 
\addplot coordinates {(1.72,21.29)(2.33,24.37)(3.66,27.3)(4.89,28.73)(6.56,29.35)(8.79,29.24)}; 
\addplot coordinates {(2.06,20.94)(2.46,24.56)(2.99,25.56)(3.57,26.23)(4.13,26.19)}; 
\legend{EASiST, wait-$k$, AlignAtt, EDAtt, InfiniSST}
\end{axis}
\end{tikzpicture}
}
\hspace{-3mm}
\subfigure[{MuST-C En$\rightarrow$Es}]{
\begin{tikzpicture}
\begin{axis}[
  easistAxis,
  xlabel=LAAL-CA (s),
  ytick={20,24,28,32},
]
\addplot coordinates {(1.45,26.61)(1.71,28.89)(2.14,30.32)(2.85,30.98)(3.9,31.86)(5.1,32.9)}; 
\addplot coordinates {(2.32,18.17)(3.05,26.81)(3.72,29.47)(4.26,30.45)(4.69,31.32)(5.15,31.93)(5.45,32.39)}; 
\addplot coordinates {(1.69,24.67)(2.34,25.81)(2.84,26.97)(3.96,30.65)(6.28,31.89)(8.37,32.95)}; 
\addplot coordinates {(1.7,25.62)(1.93,27.5)(2.78,29.4)(3.58,31.64)(4.75,33.17)(6.75,33.93)}; 
\addplot coordinates {(2.07,25.67)(2.47,29.63)(2.98,31.39)(3.52,32.23)(4.13,32.45)}; 
\legend{EASiST, wait-$k$, AlignAtt, EDAtt, InfiniSST}
\end{axis}
\end{tikzpicture}
}
\hspace{-3mm}
\subfigure[{Europarl-ST En$\rightarrow$De}]{
\begin{tikzpicture}
\begin{axis}[
  easistAxis,
  xlabel=LAAL-CA (s),
  ymax=26, xmax=10,
  ytick={14,18,22,26},
]
\addplot coordinates {(1.77,18.87)(2.41,22.12)(3.75,24.01)(5.61,25.17)(7.38,25.5)(8.39,25.45)}; 
\addplot coordinates {(2.83,13.74)(3.61,22.01)(4.49,24.61)(5.26,24.61)(5.96,25.0)(6.8,25.13)(7.42,25.26)}; 
\addplot coordinates {(2.18,13.94)(2.86,20.08)(5.0,23.9)(8.16,24.93)(11.59,24.97)(16.32,25.09)}; 
\addplot coordinates {(2.5,16.73)(3.64,20.09)(4.67,22.48)(7.77,24.48)(11.94,25.01)(15.6,25.21)}; 
\addplot coordinates {(2.56,16.05)(2.8,19.35)(3.26,21.06)(3.86,22.03)(4.45,22.54)}; 
\legend{EASiST, wait-$k$, AlignAtt, EDAtt, InfiniSST}
\end{axis}
\end{tikzpicture}
}
\hspace{-3mm}
\subfigure[{Europarl-ST En$\rightarrow$Es}]{
\begin{tikzpicture}
\begin{axis}[
  easistAxis,
  xlabel=LAAL-CA (s),
  ytick={20,24,28,32},
  xmax=9,
]
\addplot coordinates {(1.79,27.4)(2.2,30.03)(3.02,31.06)(4.39,32.34)(6.17,33.0)(7.61,33.1)}; 
\addplot coordinates {(3.04,16.38)(3.77,31.06)(4.62,32.77)(5.38,32.83)(6.05,33.02)(6.85,33.08)(7.44,33.0)}; 
\addplot coordinates {(2.26,19.23)(2.75,20.71)(4.23,24.53)(6.73,32.57)(9.49,32.91)(12.71,33.02)}; 
\addplot coordinates {(2.2,21.98)(2.94,26.49)(4.04,29.68)(6.18,32.19)(8.91,32.84)(11.54,32.93)}; 
\addplot coordinates {(2.57,22.14)(2.76,26.46)(3.17,28.48)(3.75,29.37)(4.34,29.64)}; 
\legend{EASiST, wait-$k$, AlignAtt, EDAtt, InfiniSST}
\end{axis}
\end{tikzpicture}
}

\caption{The translation quality (BLEU) against the computational-aware latency metric (LAAL-CA) on both in-domain (MuST-C En$\rightarrow$De/Es) and out-of-domain (Europarl-ST En$\rightarrow$De/Es) test sets.}
\label{fig:results_simulst_ca}
\end{figure*}

%% file: tables/speed.tex
\begin{table}[t]
\centering
\resizebox{\linewidth}{!}{%
\begin{tabular}{lccccc}
\toprule
\textbf{Method} & EASiST & wait-$k$ & AlignAtt & EDAtt & offline \\
\midrule
\textbf{Speed ($\downarrow$)} & 28.95 & 38.51 & 123.75 & 120.92 & 23.25 \\
\bottomrule
\end{tabular}
}
\caption{Comparison of inference speed (ms/token).}
\label{tab:inference_speed}
\end{table}

%% file: figures/ablation.tex
\begin{figure}[t]
\pgfplotsset{
    width=8.5cm,height=6.5cm,
    every axis y label/.append style={at={(-0.12,0.5)}},
    every axis/.append style={line width=0.9pt},
}
\centering
\begin{tikzpicture}[baseline]
\begin{axis}[
    ylabel=BLEU,
    xlabel=LAAL (s),
    enlargelimits=0.02,
    font=\small,
    label style={font=\small},
    tick label style={font=\small},
    legend style={font=\scriptsize,
    at={(0.725,0.01)},
    anchor=south,
    legend columns=1},
    legend cell align={left},
    xmajorgrids=true,
    ymajorgrids=true,
    grid style=dashed,
    ymin=17,ymax=30.5,
    xmin=1.2,
    ytick={18,20,22,24,26,28,30},
]

\addplot[color=orchidpink,mark=o, mark size=1.4pt, line width=0.9pt] coordinates {(1.26,22.67)(1.62,25.36)(2.31,27.28)(3.23,28.61)(4.19,29.32)(4.97,29.39)}; 

\addplot[color=honeyorange,mark=o, mark size=1.4pt, line width=0.9pt] coordinates {(1.33,23.54)(1.68,25.92)(2.52,27.1)(3.08,27.37)(3.56,27.41)(4.49,27.35)(5.00,26.72)}; 

\addplot[color=mintgreen,mark=o, mark size=1.4pt, line width=0.9pt] coordinates {(1.27,22.86)(1.65,25.15)(2.35,26.49)(3.29,27.26)(4.18,27.41)(5.0,27.76)}; 

\addplot[color=cyanblue,mark=o, mark size=1.4pt, line width=0.9pt] coordinates {(1.42,23.16)(1.66,24.41)(2.13,25.82)(2.89,27.25)(3.72,28.34)(4.3,28.63)}; 

\addplot[color=indigoblue,mark=o, mark size=1.4pt, line width=0.9pt] coordinates{(1.33,22.37)(1.62,24.13)(2.21,25.72)(3.09,27.31)(4.2,28.26)(4.94,28.44)}; 

\addplot[color=lavenderpurple,mark=o, mark size=1.4pt, line width=0.9pt] coordinates{(1.45,20.97)(1.75,22.18)(2.12,22.96)(2.78,23.97)(3.59,24.41)(4.12,24.74)(4.72,24.52)};

\legend{
EASiST, 
\ \ \ \ \ w/o Stage III $\mathcal{L}_{\text{policy}}$, 
\ \ \ \ \ w/o Stage III $\mathcal{L}_{\text{ST}}$, 
\ \ \ \ \ w/o Stage II, 
\ \ \ \ \ w/o Stage I, 
\ \ \ \ \ w/o Stage I and Stage II
}

\end{axis}
\end{tikzpicture}
\caption{Ablation study of our approach on the \texttt{tst-COMMON} set of MuST-C En$\rightarrow$De dataset. }
\label{fig:ablation}
\end{figure}

%% file: figures/training_time.tex
\begin{table}[t]
\centering
\small
\begin{tabular}{lcccc}
\toprule
\textbf{Setting} & \textbf{Stage I} & \textbf{Stage II} & \textbf{Stage III} & \textbf{Total} \\
\midrule
EASiST & 8 & 18 & 7 & 33 \\
w/o Stage I & 0 & 18 & 28 & 46 \\
w/o Stage II & 8 & 0 & 37 & 45 \\
w/o Stage I+II & 0 & 0 & 154 & 154 \\
\bottomrule
\end{tabular}
\caption{Training time (hours) for different ablation settings.}
\label{tab:ablation_time}
\end{table}

%% file: figures/results_ft_module.tex
\begin{figure}[t]
\pgfplotsset{
    width=8.5cm,height=6.5cm,
    every axis y label/.append style={at={(-0.12,0.5)}},
    every axis/.append style={line width=0.9pt},
}
\centering
\begin{tikzpicture}[baseline]
\begin{axis}[
    ylabel=BLEU,
    xlabel=LAAL (s),
    enlargelimits=0.05,
    font=\small,
    legend style={
    font=\scriptsize,
    at={(0.71,0.01)},
    anchor=south,
    },
    legend cell align={left},
    xmajorgrids=true,
    ymajorgrids=true,
    grid style=dashed,
    xmin=1.4,xmax=5.3,
    ymin=21,ymax=30,
]

\addplot[color=applegreen,mark=o, mark size=1.4pt, line width=0.9pt] coordinates {(1.34,22.72)(1.62,24.36)(2.18,26.25)(3.11,27.66)(4.13,28.86)(4.78,29.24)}; 

\addplot[color=cyanblue,mark=o, mark size=1.4pt, line width=0.9pt] coordinates {(1.34,22.68)(1.62,24.32)(2.16,26.29)(3.09,27.54)(4.07,28.57)(4.75,29.07)}; 

\addplot[color=indigoblue,mark=o, mark size=1.4pt, line width=0.9pt] coordinates {(1.35,22.55)(1.64,24.15)(2.18,26.11)(3.14,27.55)(4.17,28.63)(4.86,29.23)}; 

\addplot[color=lavenderpurple,mark=o, mark size=1.4pt, line width=0.9pt] coordinates {(1.34,22.62)(1.64,24.1)(2.18,25.83)(3.14,27.54)(4.15,28.58)(4.82,29.23)}; 

\addplot[color=orchidpink,mark=o, mark size=1.4pt, line width=0.9pt] coordinates{(1.26,22.67)(1.62,25.36)(2.31,27.28)(3.23,28.61)(4.19,29.32)(4.97,29.39)}; 

\addplot[color=punchpink,mark=o, mark size=1.4pt, line width=0.9pt] coordinates{(1.26,22.48)(1.71,24.84)(2.36,26.18)(3.28,27.42)(4.27,28.28)(4.93,28.39)}; 

\addplot[color=honeyorange,mark=o, mark size=1.4pt, line width=0.9pt] coordinates{(1.25,22.57)(1.64,25.11)(2.41,27.4)(3.41,28.69)(4.53,29.04)(5.26,29.32)}; 

\legend{
ALL,
Encoder + LLM,
Adapter + LLM,
LLM,
Encoder + Adapter (EASiST),
Encoder,
Adapter
}

\end{axis}
\end{tikzpicture}

\caption{BLEU-LAAL curves on En$\rightarrow$De \texttt{tst-COMMON} set when fine-tuning different modules during Stage III.}
\label{fig:trainable_module}
\end{figure}

%% file: tables/fluency.tex
\begin{table}[t]
\centering
\resizebox{\linewidth}{!}{%
\begin{tabular}{lccccccc}
\toprule
$\tau$ & 0.1 & 0.2 & 0.3 & 0.4 & 0.5 & 0.6 & offline \\
\midrule
En$\rightarrow$De & 7.51 & 7.75 & 7.96 & 8.09 & 8.18 & 8.29 & 8.39 \\
En$\rightarrow$Es & 7.84 & 8.01 & 8.07 & 8.15 & 8.21 & 8.24 & 8.32 \\
\bottomrule
\end{tabular}%
}
\caption{Fluency scores (0–10) of translations generated by EASiST under different policy thresholds ($\tau$) , evaluated by \texttt{DeepSeek-V3-0324}.} %
\label{tab:fluency}
\end{table}

%% file: sec5_conclusion.tex
\section{Conclusion}
\label{sec:conclusion}

In this work, we present \textbf{EASiST}, an efficient and adaptive framework for SimulST.  
We first introduce a novel SimulST data curation pipeline that generates monotonic, interleaved speech-translation pairs from offline corpora.  
We further introduce a lightweight policy module for adaptive read/write decisions and adopt a three-stage training strategy to progressively align text and speech modalities and optimize streaming translation performance.  
Experiments on the MuST-C and Europarl-ST En$\rightarrow$De and En$\rightarrow$Es benchmarks show that EASiST achieves better latency-quality trade-offs while maintaining high training and inference efficiency.